\documentclass{article}

\usepackage{PRIMEarxiv}

\usepackage[utf8]{inputenc} 
\usepackage[T1]{fontenc}    
\usepackage{hyperref}       
\usepackage{url}            
\usepackage{booktabs}       
\usepackage{amsfonts}       
\usepackage{nicefrac}       
\usepackage{microtype}      
\usepackage{lipsum}
\usepackage{fancyhdr}       
\usepackage{graphicx}       

\usepackage{caption}
\usepackage{subcaption}
\usepackage{authblk}

\usepackage{pdfpages}

\graphicspath{{media/}}     

\title{Optical Flow Estimation from Event-based Cameras and Spiking Neural Networks
\thanks{\textit{\underline{Citation}}: 
\textbf{Cuadrado et al. Optical flow estimation from event-based cameras and spiking neural networks. Frontiers in Neuroscience, 17, 2023 DOI:10.3389/fnins.2023.1160034.}} 
}

\author[1]{\textbf{CUADRADO Javier}}
\author[1]{\textbf{RANÇON Ulysse}}
\author[1,2]{\textbf{COTTEREAU Benoit R.}}
\author[3]{\textbf{BARRANCO Francisco}}
\author[1]{\textbf{MASQUELIER Timothée}}
\affil[1]{CerCo UMR 5549, CNRS – Université Toulouse III, Toulouse, France}
\affil[2]{IPAL, CNRS IRL 2955, Singapore, Singapore}
\affil[3]{Department of Computer Engineering, Automatics and Robotics, CITIC, University of Granada, Granada, Spain}

\begin{document}
\maketitle

\begin{abstract}
Event-based cameras are raising interest within the computer vision community. These sensors operate with asynchronous pixels, emitting events, or “spikes”, when the luminance change at a given pixel since the last event surpasses a certain threshold. Thanks to their inherent qualities, such as their low power consumption, low latency and high dynamic range, they seem particularly tailored to applications with challenging temporal constraints and safety requirements. Event-based sensors are an excellent fit for Spiking Neural Networks (SNNs), since the coupling of an asynchronous sensor with neuromorphic hardware can yield real-time systems with minimal power requirements. In this work, we seek to develop one such system, using both event sensor data from the DSEC dataset and spiking neural networks to estimate optical flow for driving scenarios. We propose a U-Net-like SNN which, after supervised training, is able to make dense optical flow estimations. To do so, we encourage both minimal norm for the error vector and minimal angle between ground-truth and predicted flow, training our model with back-propagation using a surrogate gradient. In addition, the use of 3d convolutions allows us to capture the dynamic nature of the data by increasing the temporal receptive fields. Upsampling after each decoding stage ensures that each decoder's output contributes to the final estimation. Thanks to separable convolutions, we have been able to develop a light model (when compared to competitors) that can nonetheless yield reasonably accurate optical flow estimates.
\end{abstract}


\section{Introduction}

Computer vision has become a domain of major interest, both in research and in industry. Indeed, thanks to the development of new technologies, such as autonomous vehicles or self-operating machines, algorithms able to perceive the environment have proven to be key to achieving the desired level of performance. Among the numerous visual features these algorithms can estimate, optical flow (the pattern of apparent motion on the image plane due to relative displacements between an observer and his environment) remains one of paramount importance. Indeed, this magnitude is directly linked with depth and egomotion, and its rich, highly temporal information is precious for advanced computer vision applications, e.g. for obstacle detection and avoidance in autonomous driving systems. Given the severe safety constraints associated with this kind of critical systems, accuracy and reliability are key to achieving successful models. However, achieving high levels of performance is not enough: the increasing concern about energy consumption motivates us to seek the most efficient model possible, all while retaining high-performance standards.

In the search of an energy-efficient way to estimate optical flow, we decided to focus our interest on event cameras. Unlike their regular, frame-based counterpart, this kind of sensor is composed of independent pixel processors, each  firing asynchronous events when the variation of the detected luminance since the previous event reaches a given threshold, being this event of positive polarity if the brightness has increased, and of negative polarity otherwise. This behavior translates into enormous energy savings: whereas conventional frame-based cameras are forced by design to output a frame at a fixed frequency, event cameras do not trigger any events for static visual scenes. Furthermore, they show a higher dynamic range, which allows them to avoid problems such as image artifacts (e.g. saturation after leaving a tunnel while driving), and a lower latency than regular cameras, which makes them particularly suitable for challenging, highly dynamic tasks (event sensors do not suffer from motion blur, unlike their frame-based counterparts). Nevertheless, they can also be less expressive: event cameras only provide information regarding changes in luminance, and not about the luminance itself. Furthermore, most event cameras discard color information, although some devices exist with independent firing for RGB formation at each pixel, like the Color-DAVIS346 event camera used by \cite{scheerlinck2019ced} to generate their CED Dataset. Finally, event cameras usually have lower spatial resolution than regular cameras, although recent technological developments are bridging this gap (e.g. \cite{metavision}).

In search of energy efficiency, the choice of the sensor is not enough: the optical flow prediction algorithm itself also has to be as efficient as possible to achieve our goal. That is why we have resorted to Spiking Neural Networks (SNNs) to develop our model. These bio-inspired algorithms, heavily inspired by the brain, consist of independent units (neurons), each of them with an inner membrane potential, which can be excited or inhibited by pre-synaptic connections. When their inner potential reaches a certain, predefined firing threshold, one spike is sent to the post-synaptic neurons, and the membrane potential is reset. Since energy consumption on dedicated hardware is linked to spike activity, which is usually much sparser than standard analog neural networks activations, SNNs represent a more energy-efficient alternative. Moreover, in the absence of movement, no input events would be produced and fed to the network, which in turn would not trigger any spikes, and a zero-optical flow prediction would be achieved (which is indeed the desired behaviour, since no input events can only be achieved by a lack of relative motion).

Finally, optical flow being a highly temporal task, incorporating temporal context into our vision model is key to achieving acceptable levels of performance. Two alternatives exist: using stateful units within the network (e.g. LSTMs, GRUs or taking advantage of the intrinsic memory capabilities in the case of SNNs), or explicitly handling the temporal dependencies with convolutions over consecutive frames along a temporal axis. Exploiting spiking neuron inherent temporal dynamics has proven to be an extremely challenging task to achieve, and we have therefore opted for the second alternative.

To sum up, the main contributions of this article are:
\begin{itemize}
    \item a novel angular loss, which can be used with standard MSE-like functions and which helps the network to learn an intrinsic spatial structure. To the best of our knowledge, we are the first to ever use such a function for optical flow estimation.
    \item 3d-encoding of input events over a temporal dimension, leading to increased optical flow estimation accuracy
    \item a hardware-friendly downsampling technique in the form of maximum pooling, that further improves the model's accuracy
    \item a spiking neural network which can be implemented on neuromorphic chips, therefore taking advantage of their energy efficiency.
\end{itemize}

\section{Related Work}

Ever since their introduction, event cameras have been gaining ground within the computer vision community, and increasing efforts have been made to develop computer algorithms based on event data. As such, different datasets have emerged in order to solve different kinds of computer vision problems, like the DVS128 Gesture Dataset by \cite{amir2017low} for gesture classification, or the EVIMO Dataset by \cite{burner2022evimo2} for motion segmentation and egomotion estimation. Despite this interest in event vision, the significant investment that event cameras represent for most research centers and companies has led to the development of event data simulators such as CARLA by \cite{Dosovitskiy17}, as well as algorithms to perform video-to-events conversion, like the model proposed in \cite{gehrig2020video}. While lacking the intrinsic noise event data usually presents, these artificial data can nonetheless be used to efficiently pre-train computer vision neural networks, e.g. \cite{9320311} pretraining their model for depth estimation on a synthetic set of event data.

Nonetheless, for real-world applications (e.g. gesture recognition, object detection, clustering, etc.), true event recordings are preferred because simulators are still lacking realistic event noise models. Concerning depth and/or optical flow regression, two datasets have currently established themselves as the go-to choices: the MVSEC Dataset by \cite{8288670}, and the DSEC Dataset by \cite{Gehrig21ral}. While all of these datasets have proven invaluable to develop event-based computer vision algorithms, there is still an enormous gap between event-based and image-based publicly available datasets, and many authors are still forced to develop their own. For example, \cite{cordone2022object} generated their own classification data from \cite{https://doi.org/10.48550/arxiv.2001.08499} to account for the additional ``pedestrian" class.

Most models so far have either been standard Analog Neural Networks (ANNs) like \cite{Gehrig3dv2021}, exploiting gated-recurrent units to achieve state-of the art accuracy on DSEC, or hybrid analog-spiking neural networks like \cite{9811821}, combining a spiking encoder with an additional analog encoder for grayscale images, followed by a standard ANN. Other models have tried to leverage the temporal context by feeding the network with not only the events themselves, but also information on event timestamps, like the EVFlowNet model presented in \cite{zhu2018ev}. More recently, \cite{Zhang_2022_CVPR} showed temporal information to be a key in accurately estimating both optical flow and depth, achieving top results in the MVSEC and the DSEC datasets thanks to their implementation of non-spiking leaky integrators with learnable per-channel time constants. While all of these models do indeed achieve good levels of performance on their test sets, none of them manage to take advantage of the neuromorphic-friendly nature of event data, since analog blocks or additional non-spiking information prevent a deployment on neuromorphic chips.

More interesting to this work are spiking neural networks applied to event vision, be it for depth or for optical flow estimation. As far as optical flow is concerned, it is worth citing the works of \cite{hagenaars2021self}, which achieves state-of-the-art levels of performance on the MVSEC Dataset with a fully spiking architecture. More recently, \cite{kosta2022adaptive} showed that spiking neural networks can indeed compete with their analog counterparts in terms of accuracy, showing top results both in the MVSEC and in the DSEC Dataset. Finally, \cite{zhang2023event} achieves a remarkable accuracy on the MVSEC Dataset with a U-Net-like architecture and a self-supervised learning rule. However, all of these models are not implementable on neuromorphic hardware, since they either use upsampling techniques which are incompatible with the spiking nature of these devices (e.g. bilinear upsampling), or re-inject intermediate, lower-scale analog optical flow predictions, thereby violating the spiking constraint by introducing floating point values in an otherwise binary model. In addition, the choice of a self-supervised learning rule, usually linked to a photometric loss function presented in \cite{https://doi.org/10.48550/arxiv.1608.05842}, means that optical flow estimations are only provided for pixels where events occurred, therefore creating non-dense flow maps. Looking at depth prediction though, we do find some interesting strategies for fully deployable neuromorphic models. Finally, authors in \cite{ranccon2022stereospike} presented in their StereoSpike model a fully-spiking, hardware-friendly network achieving remarkable accuracy on the MVSEC Dataset, thanks to stateless spiking neurons that have greatly inspired our work.

While we have focused on optical flow and depth predictions with event cameras, there have also been preceding works achieving top results on other computer vision tasks using event datasets and spiking neural networks. Such is the case of the works of \cite{kim2022beyond}, who performed semantic segmentation via supervised training of a SNN, or the method described in \cite{kirkland2022unsupervised} that addressed instance segmentation on event data using a biologically-plausible learning strategy.

\section{Materials and Methods}

\subsection{Training Dataset}

Our study focuses on driving scenes, and we chose the DSEC Dataset by \cite{Gehrig21ral} to train our model. Unlike previous state-of-the-art datasets, such as the Muti-Vehicle Stereo Event Camera (MVSEC) Dataset by \cite{8288670}, which provided different working scenarios (indoors/outdoors, day/night, and four possible vehicle configurations: pedestrian, motorbike, car and drone), the DSEC dataset only consists of driving scenario sequences. However, it provides higher-quality ground-truth labels, thanks to the finer processing of the LIDAR measurements. In addition, this dataset also includes masks for invalid pixels, i.e., pixels where the optical flow ground-truth is unknown. As such, our metrics have only been evaluated on the valid pixels. Furthermore, this dataset provides an open benchmark to submit the results, which we used to determine our test metrics and compare ourselves to other works.

\subsection{Input Event Representation}

\begin{figure}
    \centering
    \begin{subfigure}{0.49\textwidth}
        \centering
        \includegraphics[width=\linewidth]{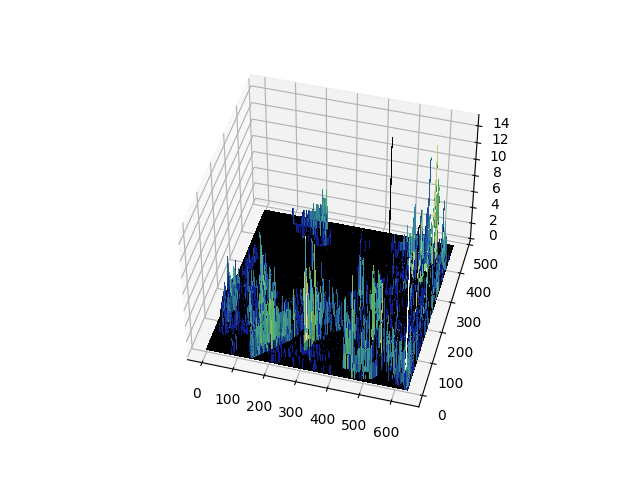}
        \caption{Perspective View}
        \label{fig:input1}
    \end{subfigure}  
    \begin{subfigure}{0.49\textwidth}
        \centering
        \includegraphics[width=\linewidth]{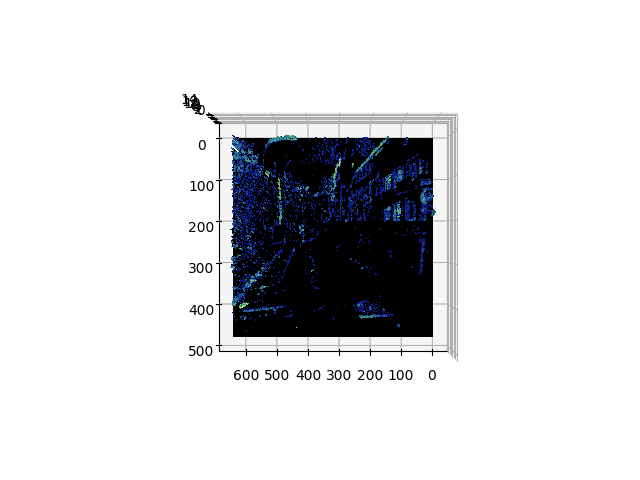}
        \caption{Top View}
        \label{fig:input2}
    \end{subfigure}  
    \caption{Example of an input frame for 1 polarity. \textbf{(A)} Event cumulation at each pixel for a given time interval. \textbf{(B)} Top view of the event frame. We can see that events are heavily linked to contours (e.g. a zebra crossing on the bottom part, or the windows on a building on the right side), while regions with constant luminance (e.g. the road or the sky) do not trigger events.}
    \label{fig:inputevents}
\end{figure}

Event cameras produce an asynchronous event $e_i$ when the luminance variation at a given pixel reaches a given threshold:
\begin{equation}
    e_i = (x_i, y_i, t_i, p_i)
\end{equation}
where $(x_i, y_i)$ are the coordinates of the pixel emitting the event, $t_i$ the event's timestamp, and $p_i$ its polarity (+1 if luminance increases, and -1 otherwise). However, in order to perform our training, we are forced to work with a discrete time model, so a pre-processing of this event stream has to be made. We therefore transform the input event stream into a sequence of frames of a given length in miliseconds, that we call ``input histograms". These frames consist of a two-channel $(C=2)$ tensor of size $(C, H, W)$, where $H$ and $W$ represent the camera's resolution, i.e., the number of input pixels and their position in the camera. At each pixel, the first channel represents the number of positive input events that have been triggered in that particular pixel during the frame's duration, and the second channel represents the number of negative events. A representation of a one-channel input frame can be found in Figure \ref{fig:inputevents}. While not a binary representation, like the representation paradigm presented in \cite{9533514}, our choice is more expressive, since event counts account for pixel relative importance and therefore provide richer spatio-temporal information.


We acknowledge that this frame-based approach increases the model's latency, since event sensors can virtually function in continuous time. However, it is imposed by the nature or our training, and is a widespread technique for event-based learning (see \cite{Gallego_2022}  on event representations). Moreover, we can leverage the latency reduction by our frame duration choice: the input stream being a continuous sequence of events, we are free to cumulate them in windows of the desired duration.

\subsection{Spiking Neuron Model}

For our network, we chose a simple neuron model that can be easily implemented with open-source Python libraries, in addition to being much less computationally expensive than closer-to-nature neuron mathematical models. This model is the \cite{mcculloch1943logical}. It was implemented using the Spikingjelly library, developed and maintained by \cite{SpikingJelly}, due to their full integration with the Pytorch library.

Our model is based on a stateless approach: the neuron's potential is reset after each forward pass. Indeed, the mathematical neuron model presented by McCulloch and Pitts consists of stateless neurons with Heaviside activation functions. This is equivalent to stateless integrate-and-fire neurons, i.e., stateless artificial neurons working as perfect integrators, but which are reset at every time step. We therefore do not exploit the intrinsic memory capabilities of spiking neurons, but rather perform a binary encoding of the information. While this approach may seem counter-intuitive, it actually further reduces energy consumption, since the reset operation is usually less energy demanding than the neuronal leak, and no resources have to be allocated to long-term memory handling. Consequently, we do not need to model such phenomenon, and as a result our neuron model is more hardware friendly than its leaky counterpart. Temporal context is handled by 3d convolutions in the encoder stages of the model, as we explain in the following section.

\subsection{Network Architecture}\label{part:archi}

\begin{figure}
\begin{center}
\includegraphics[width=0.85\linewidth]{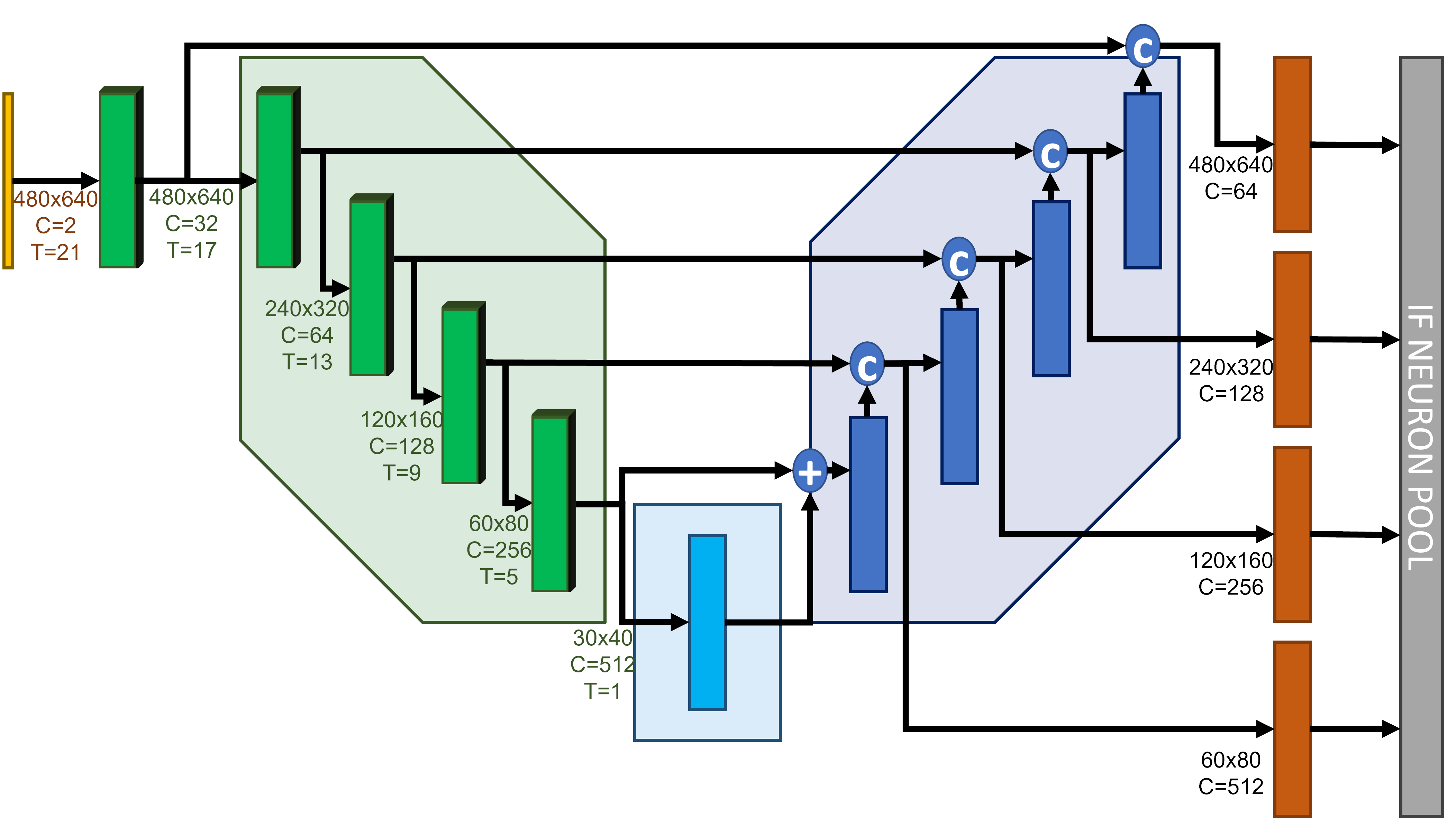}
\end{center}
\caption{Our proposed network architecture. 3D Encoders ensure the incorporation of a temporal context within the model. Downsampling is performed via max. pooling to account for spatial spike activity. Each decoding stage is upsampled to contribute to the final network prediction.}
\label{fig:archi}
\end{figure}

Our network is based on a U-Net-like architecture (\cite{ronneberger2015u}). Indeed, U-Net has established itself as a reference model when full-scale image predictions are required, i.e., predictions at roughly the same resolution as the input data. Our architecture is shown in Figure \ref{fig:archi}. After a first convolution stage which increases the number of channels to 32 without modifying the input tensor size, each encoder stage halves the tensor width and height while doubling the number of channels. Conversely, each decoder stage doubles the tensor width and height, and halves the number of channels.

In order to increase the network expressivity, each decoder stage plays a role in the final prediction. Each decoder output is upsampled into a full-scale, two-channel tensor (x- and y-components of the optical flow estimation). All of the outputs equally contribute to the network's final estimation, which consists of the combination of successive coarse predictions. The loss function is evaluated after each update of the final neuron pool, thus forcing the network's prediction to be close to the ground-truth as early as the first coarse update. This approach has been introduced in \cite{ranccon2022stereospike} and proved to be beneficial to increasing the overall accuracy.

The main features of our network are the following:

\begin{itemize}
    \item Inspired by Temporal-Convolutional Networks, presented in \cite{lea2016temporal} and \cite{lea2017temporal}, we use three-dimensional convolutions for our data encoding. Consecutive input frames are combined by the temporal kernel via unpadded convolutions, decreasing the temporal dimension in size so it collapses to 1 when reaching the bottleneck. Acting as delay lines, they allow to explicitly handle the temporal dimension. The small temporal kernel size is able to capture short-term temporal relationships, while the increasing temporal receptive field due to consecutive convolutions along the temporal dimension accounts for long-term dependencies. Afterwards, the network architecture is fully two-dimensional. By default, the temporal kernel size we use is 5, which leads to a temporal receptive field of $21 \cdot 9ms = ~189ms$ from the bottleneck and beyond.
    \item Skip connections between the encoder and the decoder consist of the last component of the temporal dimension at the corresponding encoding stage, since we believe the most recent event information to be the most relevant for optical flow estimation. We tested both sum and concatenate skip connections and found that concatenations led to the best estimations (these results are presented in subsection \ref{part:best}). 
    \item Given the relative importance of the residual blocks in the total number of parameters, and in search of the lightest possible model, we also analyzed the effect of reducing the number of residuals on the network's performance. We found that the best model only necessitated one residual, unlike other conventional U-Net-like architectures (e.g. \cite{hagenaars2021self}).
    \item Downsampling in the encoding stages is performed via maximum pooling, instead of traditional strided convolutions, to account for spikes within the kernel's region, and not so much about individual spikes. This approach has proved to increase our model's performance. To the best of our knowledge, it is the first time this technique is used in a U-Net-like spiking neural network for dense regression. In addition, \cite{gaurav2022spiking} showed that this kind of downsampling strategy is supported by neuromorphic hardware.
    \item Since our final aim is to develop a model that could be implemented on a neuromorphic chip, the whole upsampling operation is performed via Nearest Neighbor upsampling, which preserves hardware friendliness. Indeed, while other widespread techniques, such as Bilinear Upsampling, interpolate each ``pixel'', Nearest Neighbor Upsampling simply copies each value into a tensor of an increased size, without modifying it. For further illustration, a graphic representation of both upsampling techniques can be found in the Supplementary Materials (Figure S1).
    \item To further decrease the model's weight, we used depth- and point-wise separable convolutions (see e.g. \cite{chollet2017xception}) everywhere in the model. These convolutions do not only decrease the model's number of parameters, but also reduce the model's overfitting, therefore increasing its performance on unseen data.
\end{itemize}

It is important to specify that our approach is integer rather than binary-based, since some of our skip connections are additions instead of concatenations, and our bottleneck's architecture is based on tensor sums. Nevertheless, our approach remains hardware-friendly, because:
\begin{itemize}
    \item If the processing were asynchronous and event-driven, then the spikes arriving through the residual connection would typically arrive before the others. Thus, if there were two spikes, one from the residual and one from the normal connection, instead of doing an explicit ADD, both spikes could be fed through the same synapse, and each spike would cause an increment of w (instead of adding the two spikes to get 2 and then multiplying by w to get the increment). Moreover, even if the spikes arrived synchronously, they would be processed sequentially using FIFO.
    \item Concatenation is equivalent to addition as a skip connection if the weights are duplicated and kept tight. Indeed, if there were two spikes, one from the residual and one from the "normal" connection, instead of doing $2 \cdot w$, the algorithm would perform $w + w$. Since the duplicated weights would be tight, the number of trainable parameters would be the same, and both operations would be equivalent.
\end{itemize}

\subsection{Supervised Learning Method}

Our model was trained with supervised learning using the surrogate gradient descent, using a sigmoid function as our surrogate gradient model. The ground-truth optical flow values were those provided in the DSEC database. While traditional self-supervised methods restrict their optical flow processing to pixels where events occurred (e.g. \cite{hagenaars2021self}, \cite{zhu2018ev} or \cite{kosta2022adaptive}, to cite a few examples), our approach permits 
dense estimations (thanks to surrogate gradient learning). We trained our model on the valid pixels given by the dataset masks at each timestep.

Our loss function included two terms:
\begin{itemize}
    \item A standard MSE-like loss between the value of the predicted flow and its corresponding ground truth, with the following formula:
    \begin{equation}
        L_{mod} = \frac{\sum^{N_{pixels}}\sqrt{(pred_x - gt_x)^2 + (pred_y - gt_y)^2}}{N_{pixels}}
    \end{equation}
    The term $N_{pixels}$ represents the number of valid pixels to be trained at each timestep.
    \item In addition to a penalization in modulus discrepancy between the vectors, we explicitly encourage the optical flow direction to be the same between the ground-truth and the prediction. This term has proven to be key to reduce noise in optical flow predictions, since pixels with low optical flow values consistently yield small modulus loss values regardless of their direction. We used the following formula:
    \begin{equation}
        L_{ang} = \frac{\sum^{N_{pixels}}acos(c\theta)}{N_{pixels}} \quad / \quad c\theta = \frac{\vec{gt}\cdot\vec{pred} + \epsilon}{|\vec{gt}|\cdot|\vec{pred}| + \epsilon}
    \end{equation}
    where $c\theta$ is the cosine of the error angle between the predicted and the ground-truth flow, and epsilon is a small parameter ($\epsilon=10^{-7}$) to ensure that no errors are found within the code during execution. Furthermore, the values of $c\theta$ are clamped between ($-1 + \epsilon, 1 - \epsilon$) for the same reason.
\end{itemize}
The final loss function used to train the model is:
\begin{equation}
    L = \lambda_{mod} \cdot L_{mod} + \lambda_{ang} \cdot L_{ang}
\end{equation}
From preliminary tests, we found that $\lambda_{mod} = \lambda_{ang} = 1$ yields good results, and we therefore decided to use these values.

As explained in subsection \ref{part:archi} (Network Architecture), each decoder's output plays a role in the final optical flow estimation. As such, and in order to encourage accuracy since the first decoder's upsampling, the loss function is evaluated for each consecutive contribution to the final pool. After each upsampling of the decoder's output, the inner potentials of an IF layer are updated, and the loss is evaluated on those potentials equivalent to summing the spikes out of each decoder stage weighted by the corresponding intermediary prediction layer.

Finally, in order to perform the back-propagation in our supervised training method, we resorted to surrogate gradient learning, introduced in \cite{8891809}, and already implemented in the SpikingJelly library \cite{SpikingJelly}.

\subsection{Training Details}

All of our calculations were performed on either NVIDIA A40 GPUs, or in Tesla V100-SXM2-16GB GPUs belonging to the French regional public supercomputer CALMIP, owned by the Occitanie region.

Trainings were realized with a batch size of 1, since it is the optimal value we have found for our task. Although unconventional, this result is in line with the one found in \cite{ranccon2022stereospike}, where a batch size of one was found optimal for depth regression from event data using stateless spiking neurons. We used an exponential learning rate scheduler, and have implemented random horizontal flip as a data augmentation technique to improve performance. Furthermore, thanks to our stateless approach, we were able to train our network with shuffled samples, instead of being forced to use the input frames sequentially.

\section{Results}

We divided our dataset into a train and a validation split, and our performance levels are reported with regard to the validation set. The exact sequences used in each split can be found in the supplementary materials. Nevertheless, we resort to the official DSEC benchmark to compare ourselves to the state-of-the-art, since it represents an objective, third-party test set. We now proceed to present the results we obtained in our studies. Due to the number of tests that we have run, all of the corresponding plots are provided in the Supplementary Materials.

\subsection{Finding the optimal kernel size}

Convolutional neural networks have regained the interest of the deep learning community during the past few years, thanks to their ability to capture spatial relations within their kernel. Recently, increased kernel sizes have been replacing the traditional 3x3 formula, with examples as relevant as \cite{liu2022convnet}, which uses 7x7 kernels. \cite{Ding_2022_CVPR} presents a method to scale up the kernel size to 31x31, and \cite{liu2022more} goes even further and proposes to go up to 51x51 for the spatial kernel size, although both of these methods rely on sparsity and re-parametrization to achieve their goal. Starting from a naive U-Net like model, we started our research by trying to optimize our spatial kernel size. In the end, our results do match those presented in \cite{Ding_2022_CVPR}, showing that 7x7 kernels are optimal. Indeed, further increasing the kernel size makes computational time explode, while accuracy plateaus. We therefore decided to adopt a 7x7 kernel in the spatial dimension for our model.

Next, we optimized the temporal kernel size, directly linked with the number of frames that we input to our model. Since we want the temporal dimension to collapse to one in the bottleneck thanks to unstrided convolutions in the temporal dimension, a larger kernel size naturally requires a greater number of frames, and therefore a heavier model. Nonetheless, it also takes into account a longer temporal context, which may be beneficial for the network's accuracy. As such, we tested our simple model for temporal kernel sizes of 3 (11 input frames), 5 (21 input frames) and 7 (31 input frames). Our results show that increasing the kernel size up to 5 does indeed boost the model's accuracy, but going beyond this size does not translate into an accuracy improvement. Thus, a temporal kernel size of 5 was chosen for the 3d convolutions in our model.

\subsection{Finding the best network architecture}\label{part:best}

In order to find the best network architecture, we evaluated two possible options:
\begin{itemize}
    \item We compared sum vs. concatenate skip connections, since concatenate skip connections are easier to implement in neuromorphic hardware, but slightly increase the number of parameters in the network.
    \item Seeking to develop a model as light as possible, we also characterized the effect of the number of residuals in the network's bottleneck on the model's performance.
\end{itemize}

After training each of the models for 35 epochs, we found the best model to be the 1-residual network with concatenate skip connections, which amounts to a total of 1.22 million of parameters and leads to an accuracy of 1.1 pixels/second of average end-point error on our validation dataset, using 9ms frames as an input in all cases. The results regarding the architecture optimization have been summarized in Table \ref{tab:architecture comparison}.

\begin{table}
    \centering
    \begin{tabular}{c c c c}
         \textbf{Model} & \textbf{Mod Loss} & \textbf{Angular Loss} & \textbf{Num. Params (M)} \\
         \\
         \hline
         \\
         1 residual + sum skip connections & 1.18 & 0.101 & 1.1 \\
         \textbf{1 residual + cat skip connections} & \textbf{1.10} & \textbf{0.094} & \textbf{1.2} \\
         2 residual + sum skip connections & 1.15 & 0.097 & 1.7 \\
         2 residual + cat skip connections & 1.16 & 0.109 & 1.8 \\
    \end{tabular}
    \caption{Performance comparison for the different proposed architectures. All of the models have been trained for 35 epochs, using 21 input frames of 9ms each.}
    \label{tab:architecture comparison}
\end{table}

\subsection{Optimizing the frame duration}

Next, we focused our attention on the optimal frame duration to accurately estimate optical flow, i.e., the total temporal context the network processes when making a prediction. This parameter is directly linked with the latency the model can achieve, since optical flow estimations are only produced at the end of each frame (provided that the input tensors are treated as a sliding window, where only the last N=21 frames are considered).

We trained the network with frames of 4.5~ms, 9~ms and 18~ms respectively. Our results show that the optimal frame duration was 9~ms, followed by 18~ms, and finally we get the worst performance for frames of 4.5~ms. While it may seem counter-intuitive as a results, since 4.5~ms frames contain a finer representation of the event sequence, we believe this phenomenon is caused by the lack of overall temporal context. Indeed, by using short frames, the network is unable to extract longer-term dependencies, and therefore to accurately predict optical flow. That is also why we believe that 18m~s frames, while coarser, do manage to better capture these long term dependencies, and therefor provide a more accurate estimation. These results, as well as all of the successive optimization studies we have performed, can be found on Table \ref{tab:optimization}.

\begin{table}
    \centering
    \begin{tabular}{c c c c}
         \textbf{Model} & \textbf{Modifications} & \textbf{Mod Loss} & \textbf{Angular Loss} \\
         \\
         \hline
         \\
         \textbf{1 res + cat} & \textbf{-} & \textbf{1.10} & \textbf{0.094} \\
         2 res + sum & - & 1.15 & 0.097 \\
         \\
         1 res + cat & 4.5ms frames & 1.41 & 0.129 \\
         2 res + sum & 4.5ms frames & 1.42 & 0.130 \\
         1 res + cat & 18ms frames & 1.19 & \underline{0.087} \\
         2 res + sum & 18ms frames & 1.32 & 0.102 \\
         \\
         1 res + cat & combined polarities & 1.14 & 0.092 \\
         2 res + sum & combined polarities & 1.31 & 0.112 \\
         
    \end{tabular}
    \caption{Performance comparison. The two best models have been tested for different slight modifications of the architecture, keeping the number of parameters mostly unchanged.}
    \label{tab:optimization}
\end{table}

\subsection{Comparison with the state-of-the-art}

We trained our best architecture on the whole DSEC dataset for a total of 100 epochs. We evaluated our model on the official test set provided by DSEC. Results are shown on Table \ref{tab:sota}. In order to provide a fair comparison, we only included results on the official benchmark, and not those reported on custom validation sets. While still far from the best models, we demonstrate the power of spiking neural networks when applied to dense regression in computer vision, achieving good levels of performance with a fraction of parameters when compared to other models.

\begin{table}
    \centering
    \begin{tabular}{l c c c}
         \textbf{Model} & \textbf{AEE (px/s)} & \textbf{AAE (deg)} & \textbf{Num. Params (M)}\\
         \\
         \hline
         \\
         \begin{tabular}[c]{@{}l@{}}E-RAF\\ \cite{Gehrig3dv2021}\end{tabular} & \textbf{0.79} & \textbf{2.9} & 5.3 \\
         \\
         Ours & 1.71 & 6.3 & \textbf{1.2} \\
         \\
         \begin{tabular}[c]{@{}l@{}}MultiCM\\ \cite{shiba2022secrets}\end{tabular} & 3.47 & 14.0 & - \\
    \end{tabular}
    \caption{Comparison with the state-of-the art, obtained from \url{https://dsec.ifi.uzh.ch/uzh/dsec-flow-optical-flow-benchmark/}}
    \label{tab:sota}
\end{table}

\begin{figure}
    \centering
    \begin{subfigure}{0.5\textwidth}
        \centering
        \includegraphics[width=\linewidth]{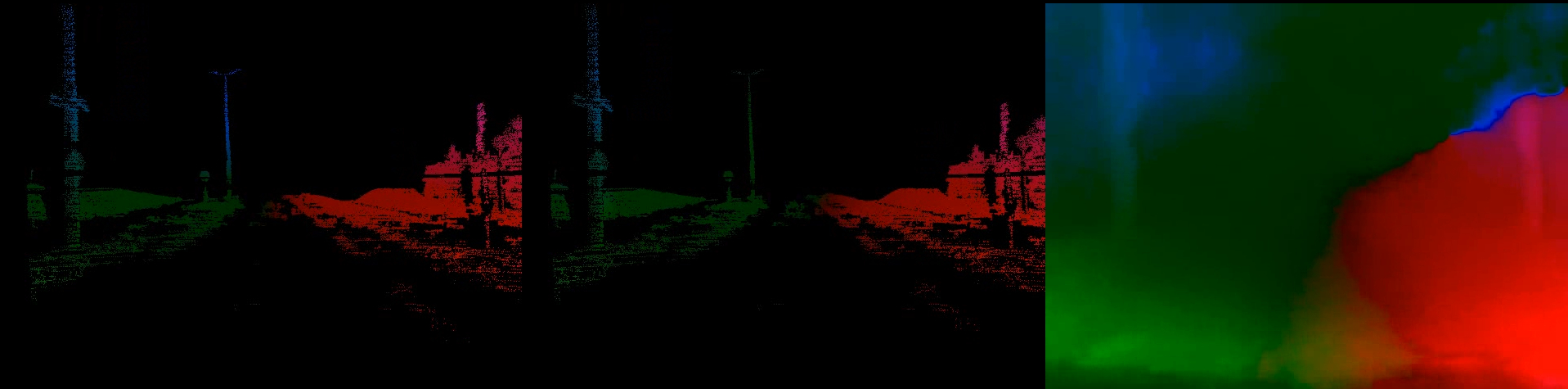}
        \caption{Optical flow discontinuities due to vertical artifacts within the visual scene.}
        \label{fig:pred1}
    \end{subfigure}  
    \begin{subfigure}{0.5\textwidth}
        \centering
        \includegraphics[width=\linewidth]{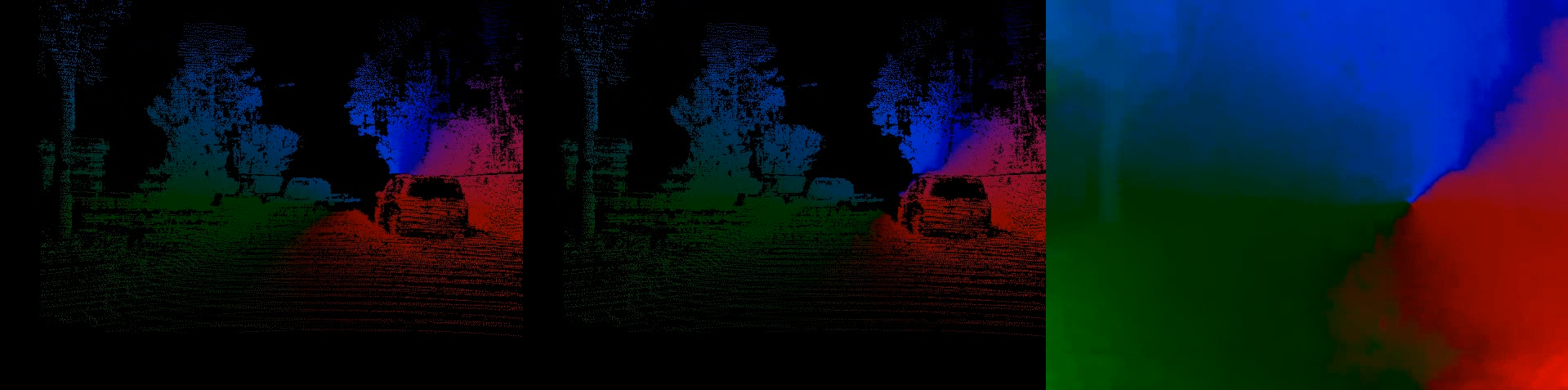}
        \caption{The silhouette of the leftmost tree can be perceived on the unmasked optical flow map.}
        \label{fig:pred5}
    \end{subfigure}
    \begin{subfigure}{0.5\textwidth}
        \centering
        \includegraphics[width=\linewidth]{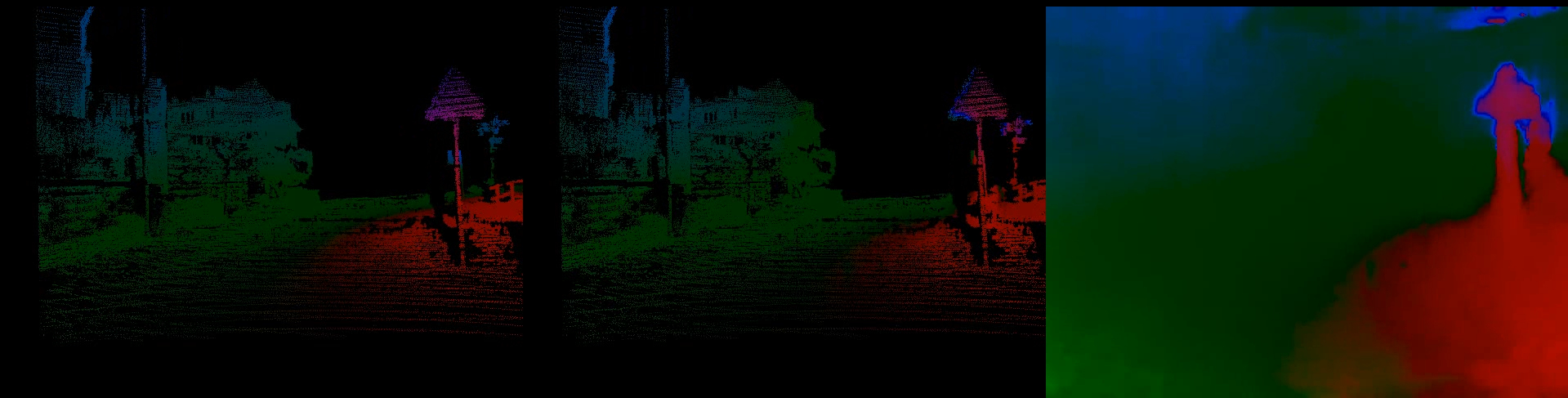}
        \caption{Traffic signs clearly distinguishable on the right side}
        \label{fig:pred6}
    \end{subfigure} 
    \begin{subfigure}{0.5\textwidth}
        \centering
        \includegraphics[width=0.5\linewidth]{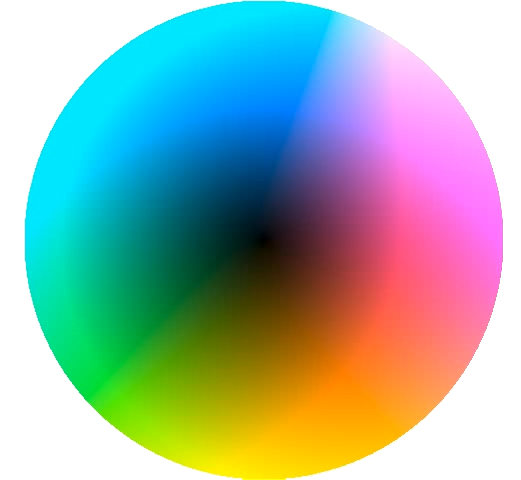}
        \caption{Optic Flow Encoding}
        \label{fig:scale}
    \end{subfigure}
    \caption{Example predictions of our best architecture on our validation set. For every picture, the leftmost image is the ground-truth, the middle image shows our masked estimation (only on valid pixels), and the rightmost image represents the unmasked estimation. (\textbf{D}) Represents the chosen colormap for optical flow representation: Optical flow is encoded as an Lab image, where the luminance channel represents the absolute magnitude of the flow, and the a and b channels the different directions.}
    \label{fig:preds}
\end{figure}

We also provide some of our model's results on the validation set, which can be found in Figure \ref{fig:preds}. These pictures show that, even if the network was not explicitly trained to distinguish image contours (since it was only trained on a selection of valid pixels at each timestep), it is nonetheless capable of extracting structural information within the scene and generalizing it, as illustrated in the rightmost images (unmasked predictions) for the given examples. These results demonstrate the model's general comprehension of the visual scene, and we believe represent a solid understanding of the pattern of motion.

\subsection{Ablation studies}

Several ablation studies have been performed on our best model to further demonstrate our claims, and we have gathered our conclusions in the following paragraphs. Plots containing all of these results can be found in the Supplementary Materials.

\subsubsection{Pooling vs. Convolutional Downsampling}

Our results show that using maximum pooling instead of strided convolutions is an efficient technique to downsample spiking data. We believe that the reason behind this behavior is that pooling is a way of densifying the tensors without changing their spiking nature.

\subsubsection{3d vs. 2d Encoding}

We also compared our baseline 3d model with an equivalent 2d model, where the 21 input frames have been fed to the network concatenated along the channel dimension, so that both models have the same temporal context. We found out that fully 2-dimensional models lead to decreased performance. We believe this is due to the fact that, by using 2d convolutions, all the temporal information is directly mixed during the first convolution stage, therefore hindering the network from finding long-term dependencies.

\subsubsection{Loss function}

We also analyzed the influence of the loss function on the final results obtained. We compared our proposed loss model to two single-term losses:
\begin{itemize}
    \item One model with only the norm of the error vector, but without the angular loss term
    \item One loss function with only a relative loss term:
    \begin{equation}
        L_{relative} = \frac{1}{N_{pixels}}\frac{\sum^{N_{pixels}}\sqrt{(pred_x - gt_x)^2 + (pred_y - gt_y)^2}}{\sqrt{gt_x^2 + gt_y^2} + \epsilon}
    \end{equation}
    This model penalizes deviations in the prediction relative to the ground truth's norm, and should therefore be able to implicitly impose a restriction on angular accuracy.
\end{itemize}

Our results show that naively limiting the error's norm is not enough to achieve competitive results, and neither is limiting the relative error. Indeed, by introducing a more aggressive term in the loss function, we managed to force the network into implicitly learning the optical flow's structure, and therefore achieve better accuracy.

It is surprising that the network with the two losses reaches a lower $L_{mod}$ than the network with $L_{mod}$ only. This shows that the second network gets trapped in a local minima and that adding the $L_{ang}$ loss helps to get out of it.

\subsubsection{Effect of combining polarities on performance}

Our next study on input representation has consisted in combining polarities into a single channel before feeding them to the model. Polarities being closely linked to phenomena like color or texture, we wish to study their influence on the final performance levels. Indeed, if we imagine a grey background with a black shape and a white shape following the same track, we would obtain opposite polarity fronts, while the optical flow pattern would be the same. We have therefore analyzed if polarities can be simply combined into a total per-pixel event count.

However, our results show that keeping separate channels for each polarities is beneficial for the network's performance. We believe this result is linked to the different dynamics linked to each of the polarities, since different thresholds lead to different behaviour for luminance increments or decrements.

\subsubsection{Skip connections in the bottleneck}

Our final ablation study targeted the very first skip connection, i.e., connecting the last encoder with the first decoder. Having always kept it as a sum (slightly redundant, given the residual block architecture) because of the high number of channels, we have also tested transforming it into a cat skip connection. However, we found out that it decreases the network's performance while also increasing the number of parameters. We therefore decided to keep it as a sum for all of the architectures.

\subsection{Model evaluation on the MVSEC Dataset}

In order to analyze the generalization capabilities of our method, we also tested our model on the Multi-Vehicle Stereo Event Camera Dataset (MVSEC), introduced in \cite{8288670}. We started by analyzing our model performance on the indoor flying sequences. To do so, we took a model pre-trained for DSEC, and optimized its weights on the MVSEC Dataset over 35 epochs. We followed a training approach akin to the one adopted for the DSEC dataset, i.e. we only considered pixels with either zero-valued ground-truth (x- and y- components of the optical flow vector below a small threshold $thr = 1e-5$) or with unknown flow values as invalid, and only trained on valid pixels. The results we obtained, as well as a comparison with other state-of-the-art models, can be found in Table \ref{tab:mvsec}. We can see that we achieve state-of-the-art performance levels on these sequences when compared to other existing spiking neural networks, and top accuracy overall, even if our architecture has not been optimized for such a vehicle/scenario configuration.

\begin{table}
    \centering
    \begin{tabular}{l c c c c}
         \textbf{Model} & \textbf{indoor\_flying1} & \textbf{indoor\_flying2} & \textbf{indoor\_flying3} & \textbf{AEE sum}\\
         \\
         \hline
         \\
         \begin{tabular}[c]{@{}l@{}}EV-FlowNet\\ \cite{zhu2018ev}\end{tabular} & 1.03 & 1.72 & 1.53 & 4.28 \\
         \\
         \cite{zhu2019unsupervised} & 0.58 & 1.02 & 0.87 & 2.47 \\
         \\
         \begin{tabular}[c]{@{}l@{}}Spike-FlowNet\\ \cite{lee2020spike}\end{tabular} & 0.84 & 1.28 & 1.11 & 3.23 \\
         \\
         \begin{tabular}[c]{@{}l@{}}Back to Event Basics$_{Evf}$\\ \cite{paredes2021back} \end{tabular} & 0.79 & 1.40 & 1.18 & 3.37 \\
         \\
         \begin{tabular}[c]{@{}l@{}}Back to Event Basics$_{Fire}$\\ \cite{paredes2021back}\end{tabular} & 0.97 & 1.67 & 1.43 & 4.07 \\
         \\
         \begin{tabular}[c]{@{}l@{}}XLIF-EV-FlowNet\\ \cite{hagenaars2021self}\end{tabular} & 0.73 & 1.45 & 1.17 & 3.35 \\
         \\
         \begin{tabular}[c]{@{}l@{}}XLIF-FireNet\\ \cite{hagenaars2021self}\end{tabular} & 0.98 & 1.82 & 1.54 & 4.34 \\
         \\
         \cite{orchard2021efficient} & 0.83 & 1.22 & 0.97 & 3.02 \\
         \\
         \begin{tabular}[c]{@{}l@{}}Fusion-FlowNet\\ \cite{9811821}\end{tabular} & \underline{0.56} & 0.95 & 0.76 & 2.27 \\
         \\
         \begin{tabular}[c]{@{}l@{}}Adaptive-SpikeNet (best ANN)\\ \cite{kosta2022adaptive}\end{tabular} & 0.84 & 1.59 & 1.36 & 3.79 \\
         \\
         \begin{tabular}[c]{@{}l@{}}Adaptive-SpikeNet (best SNN)\\ \cite{kosta2022adaptive}\end{tabular} & 0.79 & 1.37 & 1.11 & 3.27 \\
         \\
         \begin{tabular}[c]{@{}l@{}}FSFN$_{FP}$\\ \cite{apolinario2022hardware}\end{tabular} & 0.82 & 1.21 & 1.07 & 3.10 \\
         \\
         \begin{tabular}[c]{@{}l@{}}FSFN$_{HP-ADC}$\\ \cite{apolinario2022hardware}\end{tabular} & 0.85 & 1.29 & 1.13 & 3.27 \\
         \\
         \textbf{\cite{shiba2022secrets}} & \textbf{0.42} & \textbf{0.60} & \textbf{0.50} & \textbf{1.52} \\
         \\
         \underline{Ours} & 0.58 & \underline{0.72} & \underline{0.67} & \underline{1.97} \\
         \\
    \end{tabular}
    \caption{Performance comparison on the MVSEC dataset (indoor sequences), showing per-sequence and total average end-point error in pixels per second. Best result in bold, runner-up underlined. Starting from a model pre-trained on DSEC, we show state-of-the-art performance without modifying our pipeline.}
    \label{tab:mvsec}
\end{table}

Next, we also tested our model on the outdoor sequences on MVSEC: training on outdoor\_day2, and evaluation on outdoor\_day1. We present these results in Table \ref{tab:mvsec_outdoor} Although our model leads to competitive results on all of the MVSEC indoor sequences, it struggles to achieve competitive results on MVSEC outdoor sequences, both when starting from a pre-trained checkpoint or from scratch. We believe that this phenomenon is due to a combination of factors:
\begin{itemize}
    \item Our network architecture, and most precisely the spatial kernel size, has been optimized for an optical flow prediction of 480 $\times$ 640 pixels. Nevertheless, the MVSEC dataset was recorded with a different event camera, and therefore may demand a different kernel size to achieve top performance levels.
    \item Our frame duration and overall temporal context have been designed for a specific camera configuration and resolution. Again, the use of a lower resolution camera leads to different optical flow dynamics, and therefore to potentially different temporal representation.
    \item Our training procedure (learning rate, scheduler, etc.) has not been designed for such a low-resolution estimation, and therefore further optimizations are needed to increase accuracy.
    \item Finally, the outdoor\_day2 sequence of the MVSEC dataset, used for training on driving scenarios, consists of only nine minutes of recording where high frequency vibrations are constantly affecting the event camera (see \cite{zhu2018ev}). In addition, the event histograms are greatly impacted by events caused by reflections on the car dashboard. These noisy events may prevent from achieving competitive results in these sequences, since they are nonetheless responsible of inputting information to the network. In fact, only by masking that section in both the input event histogram and the associated ground-truth have we achieved training on this scenario: otherwise, the network oscillates without consistently increasing accuracy.
\end{itemize}
Nevertheless, our model achieves a certain level of learning on this condition, and we are convinced that better results could be obtained by optimizing the training pipeline for this scenario (specially the frame duration and the kernel sizes). Taking into account this learning, in conjunction with our competitive results on indoor flying scenarios, we believe that these results demonstrate the generalization capabilities of our approach, as well as its applicability in a variety of conditions.

\begin{table}[h!]
    \centering
    \begin{tabular}{l c}
         \textbf{Model} & \textbf{outdoor\_day1 (px/s)}\\
         \\
         \hline
         \\
         \begin{tabular}[c]{@{}l@{}}EV-FlowNet\\ \cite{zhu2018ev}\end{tabular} & 0.49 \\
         \\
         \underline{\cite{zhu2019unsupervised}} & \underline{0.32} \\
         \\
         \begin{tabular}[c]{@{}l@{}}\textbf{ECN$_{masked}$}\\ \textbf{\cite{ye2020unsupervised}}\end{tabular} & \textbf{0.30} \\
         \\
         \begin{tabular}[c]{@{}l@{}}Spike-FlowNet\\ \cite{lee2020spike}\end{tabular} & 0.49 \\
         \\
         \begin{tabular}[c]{@{}l@{}}Back to Event Basics$_{Evf}$\\ \cite{paredes2021back}\end{tabular} & 0.92 \\
         \\
         \begin{tabular}[c]{@{}l@{}}Back to Event Basics$_{Fire}$\\ \cite{paredes2021back}\end{tabular} & 1.06 \\
         \\
         \begin{tabular}[c]{@{}l@{}}XLIF-EV-FlowNet\\ \cite{hagenaars2021self}\end{tabular} & 0.45 \\
         \\
         \begin{tabular}[c]{@{}l@{}}XLIF-FireNet\\ \cite{hagenaars2021self}\end{tabular} & 0.54 \\
         \\
         \begin{tabular}[c]{@{}l@{}}Fusion-FlowNet\\ \cite{9811821}\end{tabular} & 0.59 \\
         \\
         \begin{tabular}[c]{@{}l@{}}Adaptive-SpikeNet (best ANN)\\ \cite{kosta2022adaptive}\end{tabular} & 0.48 \\
         \\
         \begin{tabular}[c]{@{}l@{}}Adaptive-SpikeNet (best SNN)\\ \cite{kosta2022adaptive}\end{tabular} & 0.44 \\
         \\
         \begin{tabular}[c]{@{}l@{}}FSFN$_{FP}$\\ \cite{apolinario2022hardware}\end{tabular} & 0.51 \\
         \\
         \begin{tabular}[c]{@{}l@{}}FSFN$_{HP-ADC}$\\ \cite{apolinario2022hardware}\end{tabular} & 0.48 \\
         \\
         \textbf{\cite{shiba2022secrets}} & \textbf{0.30} \\
         \\
         Ours & 0.85 \\
         \\
    \end{tabular}
    \caption{Performance comparison on the MVSEC dataset (outdoor sequences), showing average end-point error in pixels per second. Best result in bold, runner-up underlined. While far from the top performing contributions, our base pipeline is able to learn to estimate optical flow from scratch, without any optimization to make it tailored to the dataset and camera.}
    \label{tab:mvsec_outdoor}
\end{table}

\section{Discussion}

Briefly, we have presented a hardware-friendly, lightweight spiking model able to accurately estimate optical flow from event-based data collected by neuromorphic vision sensors. We propose an efficient temporal coding in the form of 3d convolutions in the encoder that increases the temporal receptive field of the deepest stages of the network. We also introduce a novel angular loss function that, in conjunction with a standard MSE-like loss, manages to boost performance by forcing the algorithm to learn the implicit spatial structure. We use maximum pooling as our downsampling strategy, thus densifying the tensors in a neuromorphic-friendly fashion. Moreover, the successive contributions of decoder outputs to the final prediction increase the network's expressivity, and allow us to achieve competitive results without resorting to intermediate prediction re-injections. Consequently, our model can be implemented in neuromorphic hardware, thus resulting in an extremely energy efficient model that can still achieve accurate predictions.

We believe our results contribute to promote spiking neural networks as energy-efficient, real-world alternatives to traditional computer vision systems, based on frame-based video treatment and/or complex sensor data. However, we acknowledge that work has yet to be done, since a lot of the intrinsic potential of SNNs, namely their inherent memory handling capabilities, has not been fully exploited in this study. Moreover, the convergence of our experiments to an optimal batch size of 1, while having indeed improved our model's performance, greatly hinders the training speed, since strategies such as data parallelization cannot be employed. We therefore believe that these results can be further improved, e.g. using techniques such as weight averaging or network pre-training.

Future research lines should focus on further combining different techniques in order to boost performance even further. For instance, exploiting the intrinsic memory of spiking neurons is indeed a potentially useful approach, but the increased computational power linked to unrolling a stateful computational graph makes the task challenging. Moreover, sensor fusion can also be explored as an alternative to boost performance, especially since most event cameras often also provide black and white images. This approach could increase the network's latency, as well as making neuromorphic implementation challenging. Furthermore, while temporal dependencies have been imposed a priori in our model, they could also be natively learnt by the network. The works of \cite{DCLS} present a way of increasing kernel sizes without an increment in network parameters, capable of achieving state-of-the-art performances. While only applied so far for 1- and 2-dimensional convolutions, their method could easily be adapted to our 3d approach.

Moreover, publicly available datasets usually lack challenging conditions, such as crossing pedestrians or vehicles, which can limit the network's generalization capabilities. While we believe that our proposed model is capable of understanding such situations (see Figure \ref{fig:pred6}, where traffic signals are easily recognizable), it would be desirable to train on more challenging scenarios.

 Finally, we would like to address hardware efficiency and implementation. We acknowledge that our approach does not provide energy savings during training, since it is performed on GPUs using standard ANN learning techniques, and therefore suffers from the same energy consumption constraints as these networks (plus the added memory usage due to the stockage of the neuron's membrane potential. However, energy savings can be achieved when deployed on dedicated hardware, since they are more energy efficient than GPUs thanks to their spiking nature. Nonetheless, even if our model is hardware-friendly, and therefore theoretically implementable on dedicated hardware, more efforts can be dedicated towards making it easier to implement. Indeed, hardware mapping would benefit from weight quantization (which would require less bits to store each synaptic weight) or sparsity encouragement to fully exploit the neuromorphic hardware advantages over GPUs. These techniques, presented by \cite{orchard2021efficient} and \cite{apolinario2022hardware}, would not only reduce energy consumption, but also facilitate potential future implementations, and should be taken into account for actual on-chip deployment.
 
\newpage

\section*{Conflict of Interest Statement}

The authors declare that the research was conducted in the absence of any commercial or financial relationships that could be construed as a potential conflict of interest.

\section*{Author Contributions}

All authors conceptualized the study and analyzed the results. JC designed, programmed, and ran the simulations. JC wrote the main core of the article. All the authors provided comments to achieve the final version of the paper.

\section*{Funding}
This research was supported in part by the Agence Nationale de la Recherche under Grant ANR-20-CE23-0004-04 ‘‘DeepSee’’, by the Spanish National Grant PID2019-109434RA-I00/ SRA (State Research Agency /10.13039/501100011033), by a FLAG-ERA funding (Joint Transnational Call 2019, project DOMINO), by the Program DesCartes and by the National Research Foundation, Prime Minister's Office, Singapore under its Campus for Research Excellence and Technological Enterprise (CREATE) Program.

\section*{Acknowledgments}
This work was granted access to the HPC resources of CALMIP supercomputing center under the allocation 2022-p22020.

The authors would also express their gratitude to the CerCo's NeuroAI team, and specially to Mr. Khalfaoui-Hassani, for their constant support and insightful feedback. We would also like to thank Mr. Fang, developer of the SpikingJelly library, for the constant support he has provided during the whole timespan of this study.

\section*{Data Availability Statement}
The DSEC Dataset, chosen to develop our models, can be found on the ``DSEC Dataset" website (\url{https://dsec.ifi.uzh.ch/}). The MVSEC dataset can be found on \url{https://daniilidis-group.github.io/mvsec/}. In addition, all of our codes can be found on the following GitHub repository: \url{https://github.com/J-Cuadrado/OF_EV_SNN}

\bibliographystyle{unsrt}  
\bibliography{references}  

\begin{thebibliography}{10}

\bibitem{scheerlinck2019ced}
Cedric Scheerlinck, Henri Rebecq, Timo Stoffregen, Nick Barnes, Robert Mahony,
  and Davide Scaramuzza.
\newblock Ced: Color event camera dataset, 2019.

\bibitem{metavision}
PROPHESEE.
\newblock Metavision® packaged sensor.
\newblock 2021.
\newblock Accessed: 2023-01-24.

\bibitem{amir2017low}
Arnon Amir, Brian Taba, David Berg, Timothy Melano, Jeffrey McKinstry, Carmelo
  Di~Nolfo, Tapan Nayak, Alexander Andreopoulos, Guillaume Garreau, Marcela
  Mendoza, et~al.
\newblock A low power, fully event-based gesture recognition system.
\newblock In {\em Proceedings of the IEEE conference on computer vision and
  pattern recognition}, pages 7243--7252, 2017.

\bibitem{burner2022evimo2}
Levi Burner, Anton Mitrokhin, Cornelia Ferm{\"u}ller, and Yiannis Aloimonos.
\newblock Evimo2: An event camera dataset for motion segmentation, optical
  flow, structure from motion, and visual inertial odometry in indoor scenes
  with monocular or stereo algorithms.
\newblock {\em arXiv preprint arXiv:2205.03467}, 2022.

\bibitem{Dosovitskiy17}
Alexey Dosovitskiy, German Ros, Felipe Codevilla, Antonio Lopez, and Vladlen
  Koltun.
\newblock {CARLA}: {An} open urban driving simulator.
\newblock In {\em Proceedings of the 1st Annual Conference on Robot Learning},
  pages 1--16, 2017.

\bibitem{gehrig2020video}
Daniel Gehrig, Mathias Gehrig, Javier Hidalgo-Carri{\'o}, and Davide
  Scaramuzza.
\newblock Video to events: Recycling video datasets for event cameras.
\newblock In {\em Proceedings of the IEEE/CVF Conference on Computer Vision and
  Pattern Recognition}, pages 3586--3595, 2020.

\bibitem{9320311}
Javier Hidalgo-Carrió, Daniel Gehrig, and Davide Scaramuzza.
\newblock Learning monocular dense depth from events.
\newblock In {\em 2020 International Conference on 3D Vision (3DV)}, pages
  534--542, 2020.

\bibitem{8288670}
Alex~Zihao Zhu, Dinesh Thakur, Tolga Özaslan, Bernd Pfrommer, Vijay Kumar, and
  Kostas Daniilidis.
\newblock The multivehicle stereo event camera dataset: An event camera dataset
  for 3d perception, 2018.

\bibitem{Gehrig21ral}
Mathias Gehrig, Willem Aarents, Daniel Gehrig, and Davide Scaramuzza.
\newblock Dsec: A stereo event camera dataset for driving scenarios, 2021.

\bibitem{cordone2022object}
Lo{\"\i}c Cordone, Beno{\^\i}t Miramond, and Philippe Thierion.
\newblock Object detection with spiking neural networks on automotive event
  data.
\newblock In {\em 2022 International Joint Conference on Neural Networks
  (IJCNN)}, pages 1--8. IEEE, 2022.

\bibitem{https://doi.org/10.48550/arxiv.2001.08499}
Pierre de~Tournemire, Davide Nitti, Etienne Perot, Davide Migliore, and Amos
  Sironi.
\newblock A large scale event-based detection dataset for automotive, 2020.

\bibitem{Gehrig3dv2021}
Mathias Gehrig, Mario Millh\"ausler, Daniel Gehrig, and Davide Scaramuzza.
\newblock E-raft: Dense optical flow from event cameras.
\newblock In {\em International Conference on 3D Vision (3DV)}, 2021.

\bibitem{9811821}
Chankyu Lee, Adarsh~Kumar Kosta, and Kaushik Roy.
\newblock Fusion-flownet: Energy-efficient optical flow estimation using sensor
  fusion and deep fused spiking-analog network architectures.
\newblock In {\em 2022 International Conference on Robotics and Automation
  (ICRA)}, pages 6504--6510, 2022.

\bibitem{zhu2018ev}
Alex~Zihao Zhu, Liangzhe Yuan, Kenneth Chaney, and Kostas Daniilidis.
\newblock Ev-flownet: Self-supervised optical flow estimation for event-based
  cameras.
\newblock {\em arXiv preprint arXiv:1802.06898}, 2018.

\bibitem{Zhang_2022_CVPR}
Kaixuan Zhang, Kaiwei Che, Jianguo Zhang, Jie Cheng, Ziyang Zhang, Qinghai Guo,
  and Luziwei Leng.
\newblock Discrete time convolution for fast event-based stereo.
\newblock In {\em Proceedings of the IEEE/CVF Conference on Computer Vision and
  Pattern Recognition (CVPR)}, pages 8676--8686, June 2022.

\bibitem{hagenaars2021self}
Jesse Hagenaars, Federico Paredes-Vall{\'e}s, and Guido De~Croon.
\newblock Self-supervised learning of event-based optical flow with spiking
  neural networks.
\newblock {\em Advances in Neural Information Processing Systems},
  34:7167--7179, 2021.

\bibitem{kosta2022adaptive}
Adarsh~Kumar Kosta and Kaushik Roy.
\newblock Adaptive-spikenet: Event-based optical flow estimation using spiking
  neural networks with learnable neuronal dynamics.
\newblock {\em arXiv preprint arXiv:2209.11741}, 2022.

\bibitem{zhang2023event}
Yisa Zhang, Hengyi Lv, Yuchen Zhao, Yang Feng, Hailong Liu, and Guoling Bi.
\newblock Event-based optical flow estimation with spatio-temporal
  backpropagation trained spiking neural network.
\newblock {\em Micromachines}, 14(1):203, 2023.

\bibitem{https://doi.org/10.48550/arxiv.1608.05842}
Jason~J. Yu, Adam~W. Harley, and Konstantinos~G. Derpanis.
\newblock Back to basics: Unsupervised learning of optical flow via brightness
  constancy and motion smoothness.
\newblock 2016.

\bibitem{ranccon2022stereospike}
Ulysse Ran{\c{c}}on, Javier Cuadrado-Anibarro, Benoit~R Cottereau, and
  Timoth{\'e}e Masquelier.
\newblock Stereospike: Depth learning with a spiking neural network.
\newblock {\em IEEE Access}, 10:127428--127439, 2022.

\bibitem{kim2022beyond}
Youngeun Kim, Joshua Chough, and Priyadarshini Panda.
\newblock Beyond classification: Directly training spiking neural networks for
  semantic segmentation.
\newblock {\em Neuromorphic Computing and Engineering}, 2(4):044015, 2022.

\bibitem{kirkland2022unsupervised}
Paul Kirkland, Davide Manna, Alex Vicente, and Gaetano Di~Caterina.
\newblock Unsupervised spiking instance segmentation on event data using stdp
  features.
\newblock {\em IEEE Transactions on Computers}, 71(11):2728--2739, 2022.

\bibitem{9533514}
Loïc Cordone, Benoît Miramond, and Sonia Ferrante.
\newblock Learning from event cameras with sparse spiking convolutional neural
  networks.
\newblock In {\em 2021 International Joint Conference on Neural Networks
  (IJCNN)}, pages 1--8, 2021.

\bibitem{Gallego_2022}
Guillermo Gallego, Tobi Delbruck, Garrick Orchard, Chiara Bartolozzi, Brian
  Taba, Andrea Censi, Stefan Leutenegger, Andrew~J. Davison, Jorg Conradt,
  Kostas Daniilidis, and Davide Scaramuzza.
\newblock Event-based vision: A survey.
\newblock {\em {IEEE} Transactions on Pattern Analysis and Machine
  Intelligence}, 44(1):154--180, jan 2022.

\bibitem{mcculloch1943logical}
Warren~S McCulloch and Walter Pitts.
\newblock A logical calculus of the ideas immanent in nervous activity.
\newblock {\em The bulletin of mathematical biophysics}, 5(4):115--133, 1943.

\bibitem{SpikingJelly}
Wei Fang, Yanqi Chen, Jianhao Ding, Ding Chen, Zhaofei Yu, Huihui Zhou,
  Timothée Masquelier, Yonghong Tian, and other contributors.
\newblock Spikingjelly.
\newblock \url{https://github.com/fangwei123456/spikingjelly}, 2020.
\newblock Accessed: 2023-01-11.

\bibitem{ronneberger2015u}
Olaf Ronneberger, Philipp Fischer, and Thomas Brox.
\newblock U-net: Convolutional networks for biomedical image segmentation.
\newblock In {\em International Conference on Medical image computing and
  computer-assisted intervention}, pages 234--241. Springer, 2015.

\bibitem{lea2016temporal}
Colin Lea, Rene Vidal, Austin Reiter, and Gregory~D Hager.
\newblock Temporal convolutional networks: A unified approach to action
  segmentation.
\newblock In {\em European conference on computer vision}, pages 47--54.
  Springer, 2016.

\bibitem{lea2017temporal}
Colin Lea, Michael~D Flynn, Rene Vidal, Austin Reiter, and Gregory~D Hager.
\newblock Temporal convolutional networks for action segmentation and
  detection.
\newblock In {\em proceedings of the IEEE Conference on Computer Vision and
  Pattern Recognition}, pages 156--165, 2017.

\bibitem{gaurav2022spiking}
Ramashish Gaurav, Bryan Tripp, and Apurva Narayan.
\newblock Spiking approximations of the maxpooling operation in deep snns.
\newblock {\em arXiv preprint arXiv:2205.07076}, 2022.

\bibitem{chollet2017xception}
Fran{\c{c}}ois Chollet.
\newblock Xception: Deep learning with depthwise separable convolutions.
\newblock In {\em Proceedings of the IEEE conference on computer vision and
  pattern recognition}, pages 1251--1258, 2017.

\bibitem{8891809}
Emre~O. Neftci, Hesham Mostafa, and Friedemann Zenke.
\newblock Surrogate gradient learning in spiking neural networks: Bringing the
  power of gradient-based optimization to spiking neural networks.
\newblock {\em IEEE Signal Processing Magazine}, 36(6):51--63, 2019.

\bibitem{liu2022convnet}
Zhuang Liu, Hanzi Mao, Chao-Yuan Wu, Christoph Feichtenhofer, Trevor Darrell,
  and Saining Xie.
\newblock A convnet for the 2020s.
\newblock In {\em Proceedings of the IEEE/CVF Conference on Computer Vision and
  Pattern Recognition}, pages 11976--11986, 2022.

\bibitem{Ding_2022_CVPR}
Xiaohan Ding, Xiangyu Zhang, Jungong Han, and Guiguang Ding.
\newblock Scaling up your kernels to 31x31: Revisiting large kernel design in
  cnns.
\newblock In {\em Proceedings of the IEEE/CVF Conference on Computer Vision and
  Pattern Recognition (CVPR)}, pages 11963--11975, June 2022.

\bibitem{liu2022more}
Shiwei Liu, Tianlong Chen, Xiaohan Chen, Xuxi Chen, Qiao Xiao, Boqian Wu,
  Mykola Pechenizkiy, Decebal Mocanu, and Zhangyang Wang.
\newblock More convnets in the 2020s: Scaling up kernels beyond 51x51 using
  sparsity.
\newblock {\em arXiv preprint arXiv:2207.03620}, 2022.

\bibitem{shiba2022secrets}
Shintaro Shiba, Yoshimitsu Aoki, and Guillermo Gallego.
\newblock Secrets of event-based optical flow.
\newblock In {\em Computer Vision--ECCV 2022: 17th European Conference, Tel
  Aviv, Israel, October 23--27, 2022, Proceedings, Part XVIII}, pages 628--645.
  Springer, 2022.

\bibitem{zhu2019unsupervised}
Alex~Zihao Zhu, Liangzhe Yuan, Kenneth Chaney, and Kostas Daniilidis.
\newblock Unsupervised event-based learning of optical flow, depth, and
  egomotion.
\newblock In {\em Proceedings of the IEEE/CVF Conference on Computer Vision and
  Pattern Recognition}, pages 989--997, 2019.

\bibitem{lee2020spike}
Chankyu Lee, Adarsh~Kumar Kosta, Alex~Zihao Zhu, Kenneth Chaney, Kostas
  Daniilidis, and Kaushik Roy.
\newblock Spike-flownet: event-based optical flow estimation with
  energy-efficient hybrid neural networks.
\newblock In {\em Computer Vision--ECCV 2020: 16th European Conference,
  Glasgow, UK, August 23--28, 2020, Proceedings, Part XXIX 16}, pages 366--382.
  Springer, 2020.

\bibitem{paredes2021back}
Federico Paredes-Vall{\'e}s and Guido~CHE de~Croon.
\newblock Back to event basics: Self-supervised learning of image
  reconstruction for event cameras via photometric constancy.
\newblock In {\em Proceedings of the IEEE/CVF Conference on Computer Vision and
  Pattern Recognition}, pages 3446--3455, 2021.

\bibitem{orchard2021efficient}
Garrick Orchard, E~Paxon Frady, Daniel Ben~Dayan Rubin, Sophia Sanborn,
  Sumit~Bam Shrestha, Friedrich~T Sommer, and Mike Davies.
\newblock Efficient neuromorphic signal processing with loihi 2.
\newblock In {\em 2021 IEEE Workshop on Signal Processing Systems (SiPS)},
  pages 254--259. IEEE, 2021.

\bibitem{apolinario2022hardware}
Marco Paul~E Apolinario, Adarsh~Kumar Kosta, Utkarsh Saxena, and Kaushik Roy.
\newblock Hardware/software co-design with adc-less in-memory computing
  hardware for spiking neural networks.
\newblock {\em arXiv preprint arXiv:2211.02167}, 2022.

\bibitem{ye2020unsupervised}
Chengxi Ye, Anton Mitrokhin, Cornelia Ferm{\"u}ller, James~A Yorke, and Yiannis
  Aloimonos.
\newblock Unsupervised learning of dense optical flow, depth and egomotion with
  event-based sensors.
\newblock In {\em 2020 IEEE/RSJ International Conference on Intelligent Robots
  and Systems (IROS)}, pages 5831--5838. IEEE, 2020.

\bibitem{DCLS}
Ismail Khalfaoui-Hassani, Thomas Pellegrini, and Timothée Masquelier.
\newblock Dilated convolution with learnable spacings.
\newblock 2021.

\end{thebibliography}



\section*{Supplementary material}

# Supplementary Material

## 1 Supplementary Data

### 1.1 Upsampling techniques illustration

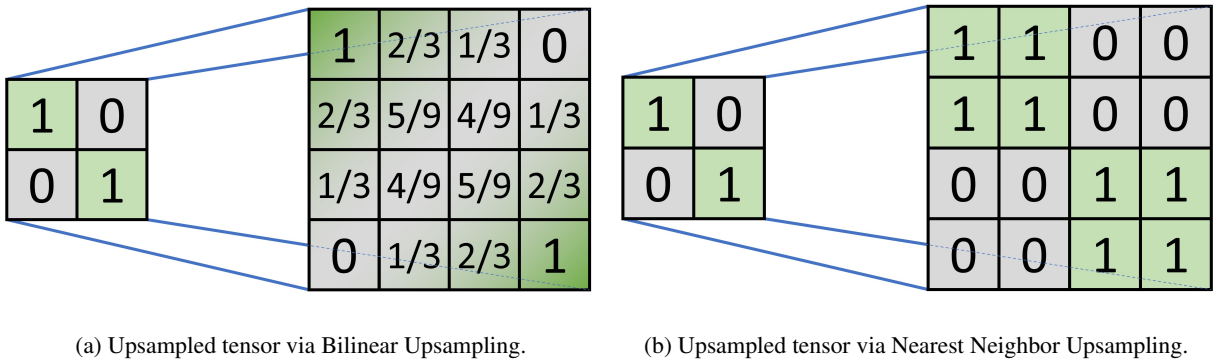

(a) Upsampled tensor via Bilinear Upsampling.      (b) Upsampled tensor via Nearest Neighbor Upsampling.

Figure 1: Upsampling techniques comparison. Unlike Bilinear Upsampling, Nearest Neighbor Upsampling guarantees binary tensors after the upsampling operation, being therefore implementable on neuromorphic hardware.

### 1.2 Train and validation split

Here is the sequence division we have done for our train and validation split:

- Trainin split:

    - zurich_city_01_a
    - zurich_city_02_a
    - zurich_city_02_c
    - zurich_city_01_e
    - zurich_city_05_a
    - zurich_city_05_b
    - zurich_city_06_a

    - zurich_city_07_a
    - zurich_city_09_a
    - zurich_city_10_a
    - zurich_city_10_b
    - zurich_city_11_a
    - zurich_city_11_c

- Validation split:

    - thun_00_a
    - zurich_city_02_d
    - zurich_city_03_a

    - zurich_city_08_a

    - zurich_city_11_b

The data split has been performed in order to ensure around 75% of the available data beeing used during training, while the remaining 25% was used in evaluation.

## 2 Supplementary Figures — Result Plots

Here are the plots containing the results of our trainings, evaluated on our validation split. Each plot consists of the modification of a single parameter or architectural block, for better comparison of their effect on performance.

### 2.1 Architecture optimization

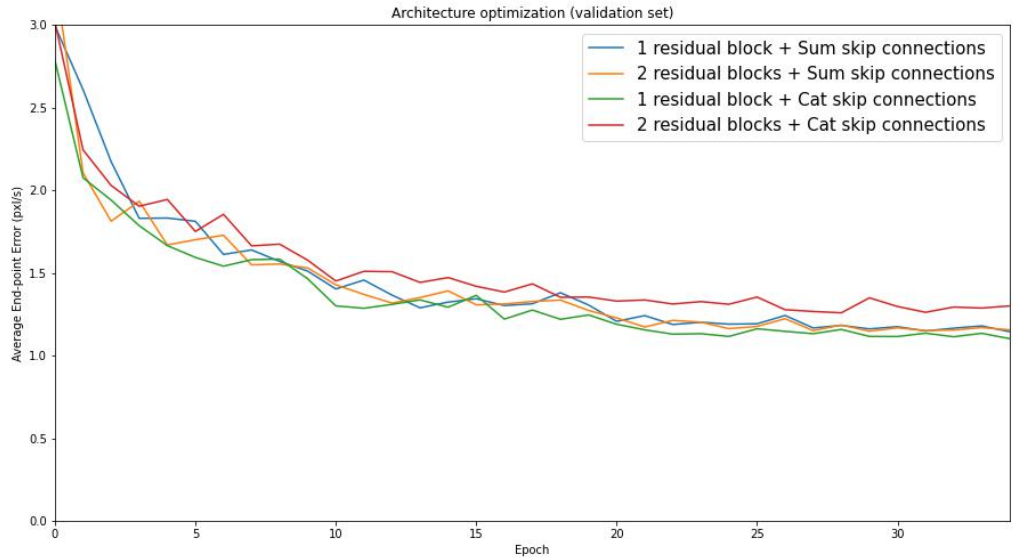

Figure 2: Architecture optimization: 1 vs. 2 residual blocks in the bottleneck,, and Sum vs. Cat skip connections. We have found the best architecture to consist of CAT skip connections and a single residual block, which amount to a total of 1.2 million parameters for our base model.

## 2.2 Frame duration optimization

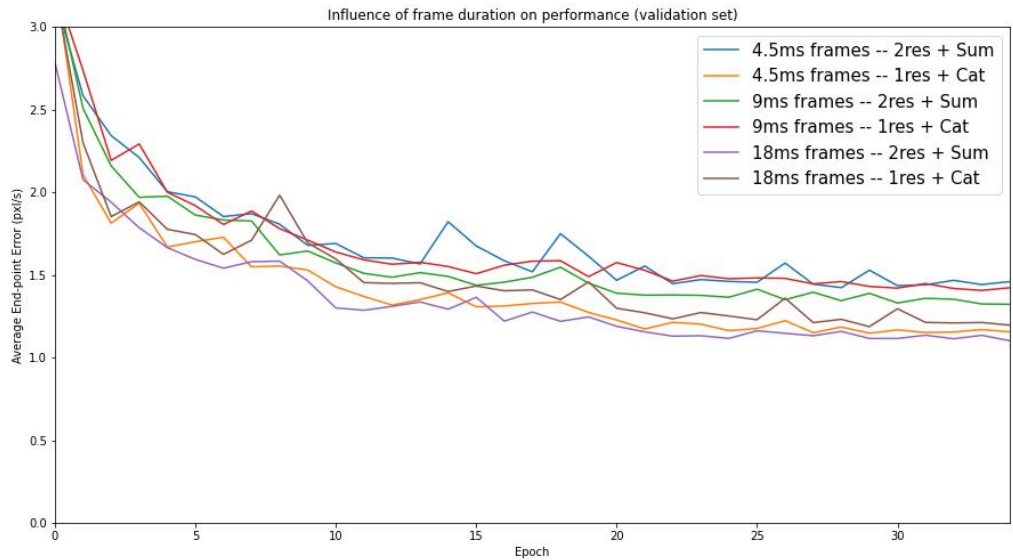

Figure 3: Frame duration optimization: 4.5ms, 9ms and 18 ms. We find intermediate frames to be the optimal choice when it comes to accuracy, since shorter histograms do not manage to capture sufficiently long-term time features, and longer histograms are not sufficiently crisp.

## 2.3 Ablation Studies

Below can be found the plots corresponding to the different ablation studies we carried out on our best model:

- Pooling vs. convolutional downsampling
- 3d vs. 2d encoding
- Loss function
- Effect of combining polarities
- Skip connections in the bottleneck

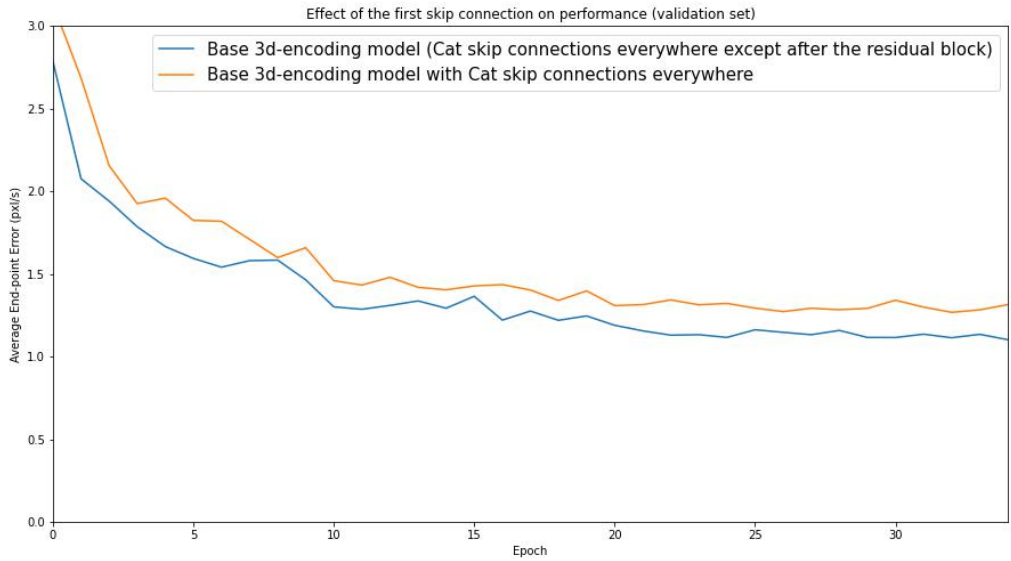

Figure 4: Accuracy comparison between strided convolutions and strided maximum pooling as downsamplnig strategies. We show that our approach, where individual spikes carry less importance (one single spike within the kernel region is enough to forward the information) achieves remarkably better results.

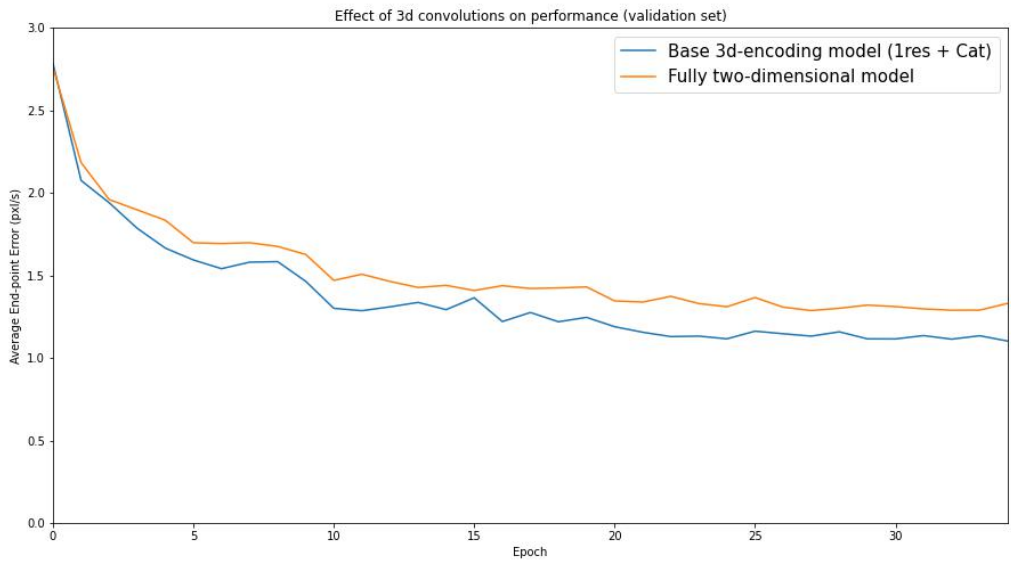

Figure 5: Accuracy comparison between our proposed 3d-encoder architecture and an equivalent fully 2-dimensional architecture. We show that explicitly handling the temporal dimension with consecutive convolutions along a temporal axis yields better quality optical flow estimations.

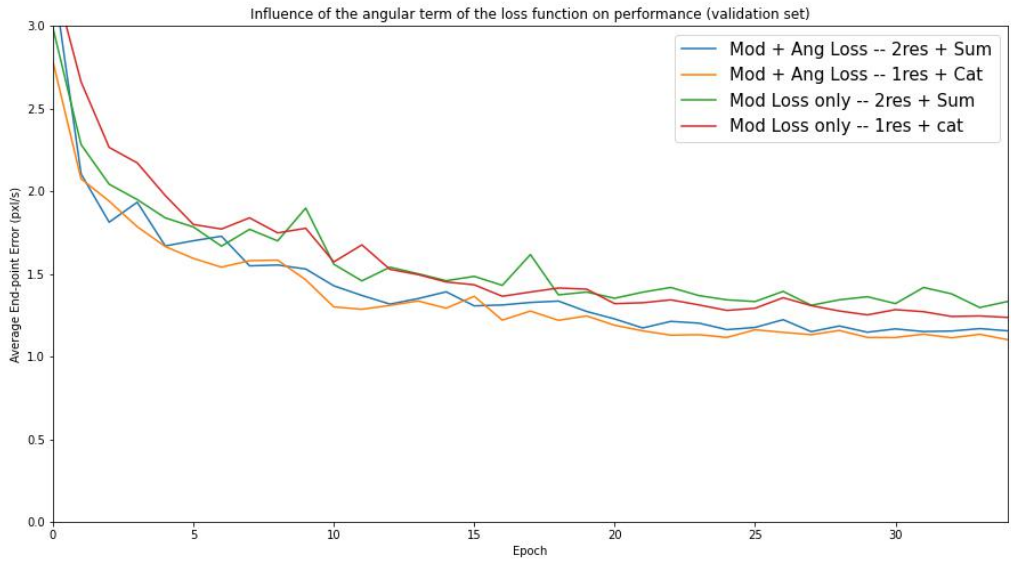

Figure 6: Influence of the angular term in the loss function on the achieved accuracy. We can observe that enforcing this term on the loss evaluation helps the neuron acquire a better general understanding of the scene, and therefore achieve better quality estimations.

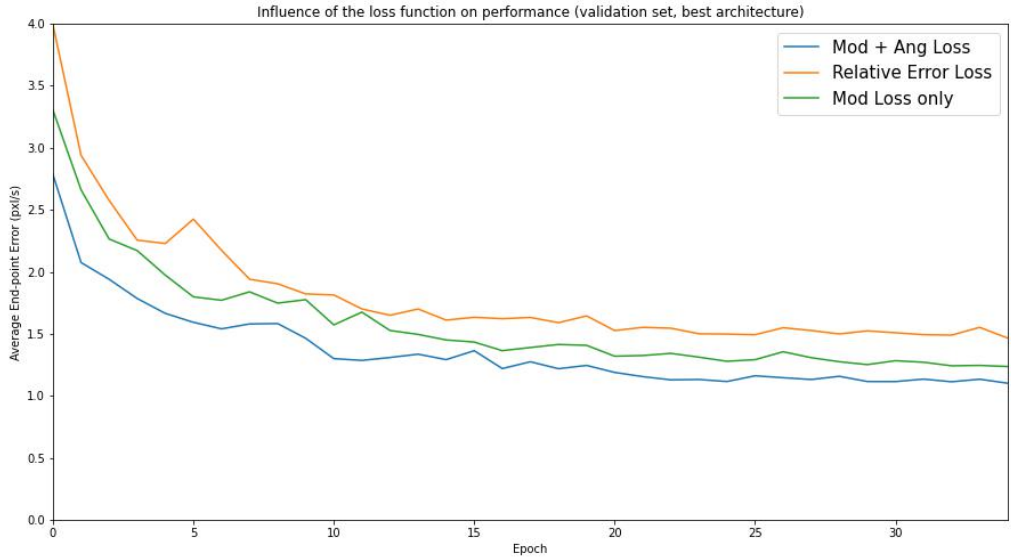

Figure 7: Influence of the loss function on the achieved accuracy (validation set): best loss model (Mod + Ang), norm of the error vector (only error norm) and relative error (error norm divided by the optical flow ground-truth magnitude).

## 3   Code availability

All of the codes we developed to carry out this study can be found on the GitHub repository `https://github.com/J-Cuadrado/OF_EV_SNN`.

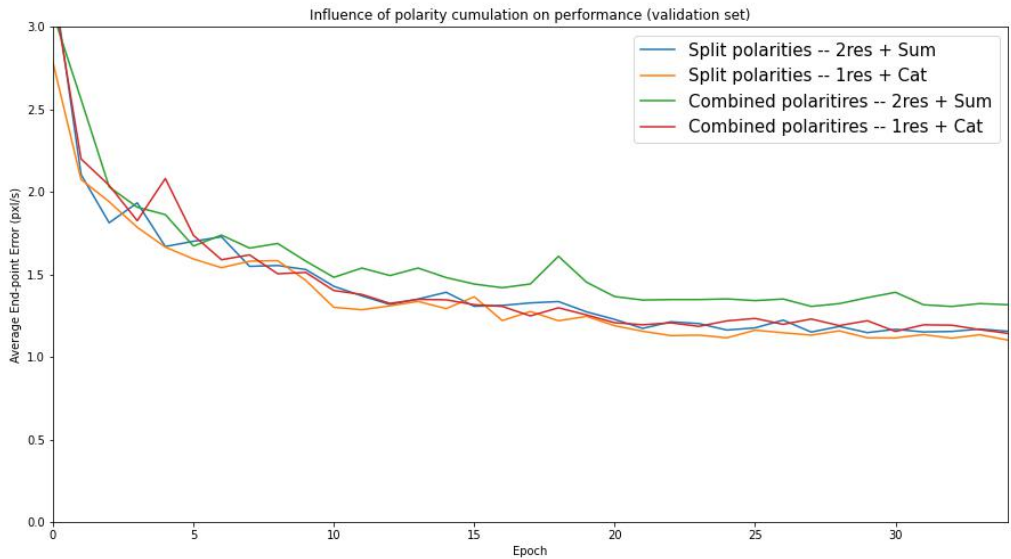

Figure 8: Effect of combining the polarities in a single channel on the final validation performance. We can observe that keeping split polarities yields better performance, although not very significantly in the case of our best architecture.

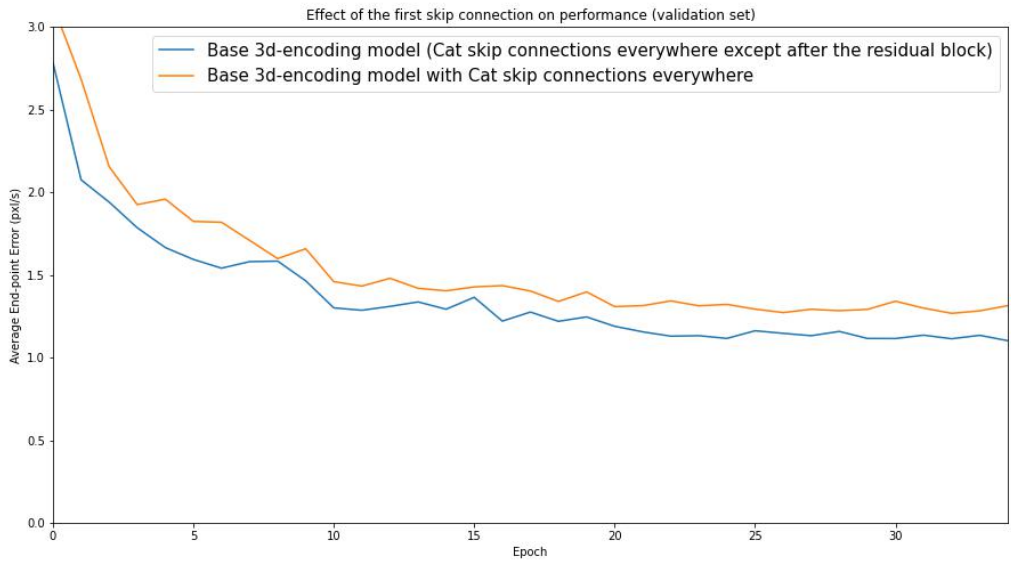

Figure 9: Influence of the first skip connection (last encoder with fisrt decoder) on the model's performance: the blue curve represents our base model, whereas the yellow curve represents a CAT skip connection between linking the encoder's output and the decoder's input.



\end{document}